%% file: ARTICLE.tex
\documentclass[letterpaper, 10 pt, conference]{ieeeconf}

\IEEEoverridecommandlockouts
\usepackage[english]{babel}
\usepackage{amsmath} 
\usepackage{amssymb}  
\usepackage{graphicx}
\usepackage{subfigure}
\usepackage{caption}
\usepackage{multirow}
\usepackage{amsfonts}
\usepackage{tabularx}
\usepackage{algorithm,algpseudocode}
\usepackage{bm}
\usepackage{enumerate}
\usepackage{verbatim}
\usepackage{float}
\usepackage{siunitx}
\usepackage{mdframed}
\usepackage{adjustbox}
\usepackage{authblk}
\usepackage{color}
\usepackage{here} 
\usepackage{physics}
\usepackage{soul}
\usepackage{xurl}
\usepackage{tabularx}
\usepackage{tikz}
\usepackage{ dsfont }
\usetikzlibrary{shapes.geometric, arrows}
\makeatletter
\let\NAT@parse\undefined
\makeatother
\usepackage{hyperref}
\usepackage{cleveref}

\usepackage{array}
\input{notation}

\input{custom_function}
\usepackage{cite}

\usepackage{wasysym}  
\author{Guanrui Li and Giuseppe Loianno
\thanks{The authors are with the New York University, Tandon School of Engineering, Brooklyn, NY 11201, USA. {\tt\footnotesize email: \{lguanrui,  loiannog\}@nyu.edu}.}
\thanks{The authors acknowledge Manling Li, Kelsey Fontenot for their help and support on this research and experiments.}
\thanks{This work was supported by the NSF CPS Grant CNS-2121391, the NSF CAREER Award 2145277, the DARPA YFA Grant D22AP00156-00, Qualcomm Research, Nokia, and NYU Wireless.}
}

\title{\LARGE \bf Nonlinear Model Predictive Control for Cooperative Transportation and Manipulation of Cable Suspended Payloads with Multiple Quadrotors}

\begin{document}

\maketitle
\thispagestyle{empty}
\pagestyle{empty}

\begin{abstract}
Autonomous Micro Aerial Vehicles (MAVs) such as quadrotors equipped with manipulation mechanisms have the potential to assist humans in tasks such as construction and package delivery. Cables are a promising option for manipulation mechanisms due to their low weight, low cost, and simple design. However, designing control and planning strategies for cable mechanisms presents challenges due to indirect load actuation, nonlinear configuration space, and highly coupled system dynamics. In this paper, we propose a novel Nonlinear Model Predictive Control (NMPC) method that enables a team of quadrotors to manipulate a rigid-body payload in all 6 degrees of freedom via suspended cables. Our approach can concurrently exploit, as part of the receding horizon optimization, the available mechanical system redundancies to perform additional tasks such as inter-robot separation and obstacle avoidance while respecting payload dynamics and actuator constraints. To address real-time computational requirements and scalability, we employ a lightweight state vector parametrization that includes only payload states in all six degrees of freedom. This also enables the planning of trajectories on the $SE(3)$ manifold load configuration space, thereby also reducing planning complexity. We validate the proposed approach through simulation and real-world experiments.
\end{abstract}
\section*{Supplementary material}
\textbf{Video}: \url{https://youtu.be/NQeEvzEkGVw}
\IEEEpeerreviewmaketitle
\input{sections/01-Intro.tex}
\input{sections/02-Model_MPC.tex}

\input{sections/03-NMPCFormulation.tex}

\input{sections/04-Experiments.tex}

\input{sections/05-Conclusion.tex}

\addtolength{\textheight}{-10.0cm}   


\bibliographystyle{IEEEtran}	
\bibliography{references}
\end{document}

%% file: notation.tex
\newcommand{\id}{k}

\newcommand{\veca}{\boldsymbol{a}}

\newcommand{\vecb}{\mathbf{b}}

\newcommand{\vece}{\mathbf{e}}
\newcommand{\vecf}{\mathbf{f}}

\newcommand{\vecg}{\mathbf{g}}
\newcommand{\vecp}{\mathbf{p}}
\newcommand{\vecq}{\mathbf{q}}

\newcommand{\vecu}{\mathbf{u}}

\newcommand{\vecx}{\mathbf{x}}

\newcommand{\vecmu}{\bm{\mu}}
\newcommand{\vecomega}{\bm{\omega}}
\newcommand{\vecrho}[1]{\bm{\rho}_{#1}}
\newcommand{\veclamb}{\bm{\Lambda}}
\newcommand{\hatvecrho}[1]{\hat{\bm{\rho}}_{#1}}
\newcommand{\vecxi}{\bm{\xi}}

\newcommand{\matG}{\mathbf{G}}
\newcommand{\matI}{\mathbf{I}}
\newcommand{\matJ}{\mathbf{J}}
\newcommand{\matK}{\mathbf{K}}
\newcommand{\matM}{\mathbf{M}}
\newcommand{\matN}{\mathbf{N}}
\newcommand{\matQ}{\mathbf{Q}}
\newcommand{\matP}{\mathbf{P}}
\newcommand{\matR}{\bm{\mathit{R}}}
\newcommand{\matU}{\mathbf{U}}
\newcommand{\matV}{\mathbf{V}}
\newcommand{\matW}{\mathbf{W}}
\newcommand{\matX}{\mathbf{X}}

\newcommand{\inertia}{\mathbf{J}}
\newcommand{\inertiaload}{\mathbf{J}_{L}}
\newcommand{\loadmass}{m_{L}}
\newcommand{\half}{\frac{1}{2}}

\newcommand{\angvel}{\mathbf{\Omega}}
\newcommand{\angacc}{\dot{\angvel}}

\newcommand{\robotpos}[1]{\vecx_{#1}}
\newcommand{\attpos}[1]{\vecp_{#1}}
\newcommand{\robotrot}[1]{\matR_{#1}}

\newcommand{\robotvel}[1]{\dot{\vecx}_{#1}}
\newcommand{\robotacc}[1]{\ddot{\vecx}_{#1}}
\newcommand{\robotangvel}[1]{\angvel_{#1}}

\newcommand{\tension}[1]{\vecmu_{#1}}
\newcommand{\tensiondes}[1]{\vecmu_{#1,des}}
\newcommand{\loadpos}{\vecx_{L}}
\newcommand{\loadquat}{\vecq_{L}}
\newcommand{\loadquatdot}{\dot{\vecq}_{L}}
\newcommand{\loadposdes}{\vecx_{L,des}}
\newcommand{\loadrot}{\matR_{L}}
\newcommand{\loadrotdes}{\matR_{L,des}}

\newcommand{\loadvel}{\dot{\vecx}_{L}}
\newcommand{\loadveldes}{\dot{\vecx}_{L,des}}
\newcommand{\loadacc}{\ddot{\vecx}_{L}}
\newcommand{\loadaccdes}{\ddot{\vecx}_{L,des}}
\newcommand{\loadangvel}{\angvel_{L}}
\newcommand{\loadangveldes}{\angvel_{L,des}}
\newcommand{\loadangacc}{\angacc_{L}}
\newcommand{\loadangaccdes}{\angacc_{L,des}}

\newcommand{\cablevec}[1]{\vecxi_{#1}}

\newcommand{\cablevel}[1]{\vecomega_{#1}}
\newcommand{\cableveldes}[1]{\vecomega_{#1,des}}
\newcommand{\cabledotvec}[1]{\dot{\vecxi}_{#1}}

\newcommand{\cableddotvec}[1]{\ddot{\vecxi}_{#1}}

\newcommand{\inputforce}[1]{\vecu_{\mathit{#1}}}
\newcommand{\inputperp}[1]{\vecu_{\mathit{#1}}^{\perp}}
\newcommand{\inputpara}[1]{\vecu_{\mathit{#1}}^{\lVert}}

\newcommand{\aptpos}[1]{\vecp_{#1}}
\newcommand{\aptvel}[1]{\dot{\vecp}_{#1}}

\newcommand{\realnum}[1]{\mathbb{R}^{#1}}
\newcommand{\SOthree}{SO(3)}

\newcommand{\twonorm}[1]{\left\lVert#1\right\rVert_2}
\newcommand{\prths}[1]{\left(#1\right)}

\newcommand{\worldf}{\mathcal{I}}
\newcommand{\robotf}[1]{\mathcal{B}_{#1}}
\newcommand{\loadf}{\mathcal{L}}
\newcommand{\axis}[2]{\mathbf{e}_{#1}^{#2}}

\newcommand{\netplforce}{\mathbf{F}}
\newcommand{\netplforcedes}{\mathbf{F}_{des}}
\newcommand{\netplmoment}{\mathbf{M}}
\newcommand{\netplmomentdes}{\mathbf{M}_{des}}

%% file: sections/01-Intro.tex
\section{Introduction}\label{sec:intro}
Low-cost autonomous Micro Aerial Vehicles (MAVs) equipped with manipulation mechanisms hold great potential for assisting humans with complex and hazardous tasks, including construction, delivery, and inspection. In construction scenarios, a team of MAVs can work collaboratively to transport construction materials from the ground to the upper floors, thereby expediting the construction process. Similarly, in urban settings where ground traffic can cause delays during rush hour, MAVs can utilize unobstructed airspace to facilitate prompt humanitarian missions or emergency medical care deliveries. These tasks require aerial robots to possess the capability to transport and manipulate objects.

\begin{figure}[t!]
\centering
  \includegraphics[width=0.8\columnwidth]{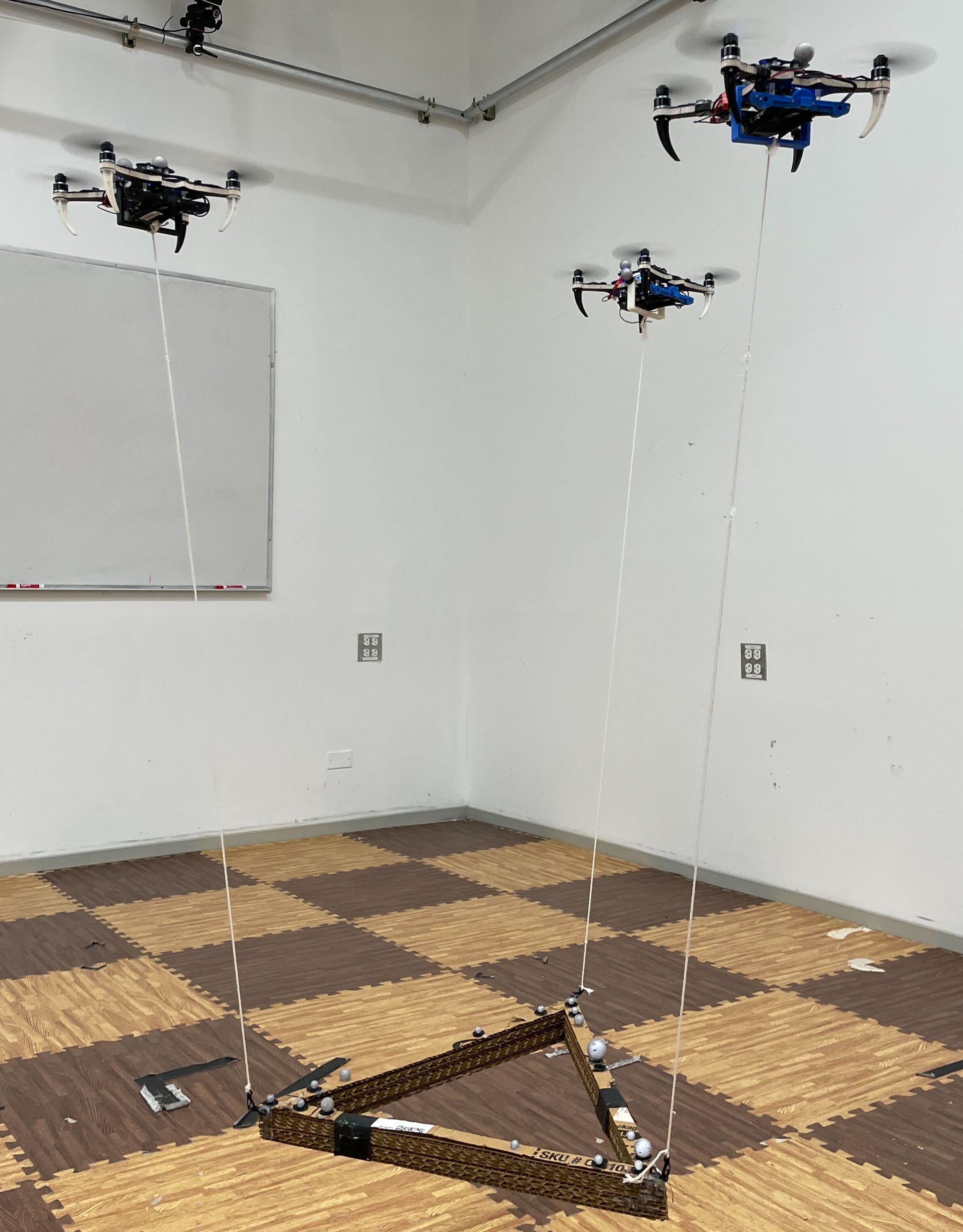}  
  \caption{Three quadrotors carrying a rigid-body payload via cables using the proposed NMPC. \label{fig:intro}}
  \vspace{-20pt}
\end{figure}
We can use various mechanisms to transport and manipulate a payload via a team of aerial robots, including ball joints~\cite{tagliabue2019ijrr,Loiannodesert2018}, robot arms~\cite{jongseok2020icra,ouyang2021aerialmanipulator}, and cables~\cite{guanrui2021iser,klausen2020passiveoutdoortransportation}, as discussed in~\cite{ollero2021TRO}. Cable mechanisms are advantageous due to their lighter weight, lower cost, simpler design requirements, and zero-actuation-energy consumption, as discussed in~\cite{guanrui2021ral}, making them particularly suitable for Size, Weight, and Power (SWaP) aerial platforms. 

Despite their benefits, these systems pose unique challenges for designing control and planning strategies due to their indirect load actuation and complex, nonlinear configuration space, and highly coupled nonlinear system dynamics between the transported load and the aerial robots. Therefore, the design of effective control and planning techniques for these systems can contribute to mitigating these challenges. In this paper, we address the challenge of designing an NMPC  that enables a team of quadrotors to manipulate and transport a rigid-body load in $6$ Degrees of Freedom (DoF) via suspended cables.

In some cases, it is sufficient to consider the load as a disturbance to the quadrotor team formation~\cite{klausen2020passiveoutdoortransportation, liang_fault-tolerant_2021}. Each quadrotor leverages a disturbance estimator to estimate the disturbance introduced by the payload. Further, they add a feedforward term in the proposed control input to counteract the disturbance and maintain team formation. A similar approach is presented in a leader-follower fashion, where the leader robot navigates the system to the goal location. The followers stabilize the cable direction around the equilibrium state as in~\cite{pratik2020icra, heng2022access} or adapt the external disturbance from the payload to augment their planned trajectory~\cite{tagliabue2017icracollaborativetransportation}. Although such methods avoid modeling the complex nonlinear system dynamics of the payload, they are conservative on motion as they try to avoid or minimize the swing of the cables. In addition, such methods can only transport the payload in $3$ DoF but not manipulate in $6$ DoF. Other techniques choose to model the coupled system dynamics and propose corresponding controllers to manipulate the payload~\cite{lee2018tcst, wu2014geometric,fink2011ijrr,Michael2011,lee2013cdccooperativepointmass}. For instance, several works~\cite{lee2013cdccooperativepointmass, Jackson_Howell_Shah_Schwager_Manchester_2020} simplify the payload's model as a point mass and designs the corresponding controller or planner to transport the payload. However, these proposed controllers cannot manipulate the payload in $6$ DoF like the aforementioned works. 

On the other hand, researchers propose to consider the carried payload in the system as a rigid body. For example, several works~\cite{fink2011ijrr,Michael2011} model the system mechanics when the robot and payload are at an equilibrium state. Subsequently, they propose a payload pose controller that allows the robots to manipulate the payload to the desired pose assuming quasi-static motion. However, such a method shares similar disadvantages to the aforementioned works  as they have conservative quasi-static motion assumptions on the payload. In~\cite{RSS2013Koushil, wu2014geometric,lee2018tcst}, researchers model the full complex nonlinear system dynamics in the system using Lagrangian mechanics. Based on this model, \cite{wu2014geometric,lee2018tcst} propose nonlinear geometric controllers for the quadrotor team to manipulate and transport a rigid-body payload to track a trajectory in full $6$ DoF and examine the method in simulation. In our recent work, we design a hierarchical formulation of the nonlinear geometric controllers and examine them in real-world experiments~\cite{guanrui2021iser,guanrui2021ral,guanrui2022rotortm}. 

As previously discussed in~\cite{RSS2013Koushil,lee2018tcst}, when the team carrying a load comprises more than three quadrotors, redundancies arise in the cable tension forces that act on the rigid-body payload. These redundancies can be exploited to accomplish secondary tasks such as obstacle avoidance, in addition to manipulation. However, previous works only considered the minimum norm solution for distributing the cable tension forces and did not explore these secondary tasks. In our recent work,~\cite{guanrui2022humanrobot}, we present two methods to exploit the redundancy for any $n\geq3$ quadrotors carrying a rigid-body payload, enabling additional tasks like obstacle avoidance and inter-robot spatial separation while still allowing payload manipulation in all $6$ DoF. However, all the previously presented works did not address the problem of satisfying additional constraints (e.g., actuator or perception constraints) during these tasks.
Recent works~\cite{geng2022jais, gengJGCD2020} propose to utilize the control redundancies in cable tension force space to achieve inter-robot spatial separation and respect the actuator constraints. However, these works are designed specifically for a team of four robots. 

Therefore, more recently, researchers have applied optimal control methods to this problem~\cite{sundin2022icradecentralizedmpctransportation,Wehbeh_Rahman_Sharf_2020_iros_nmpc,sun2023nmpcmanipulation}. In~\cite{sundin2022icradecentralizedmpctransportation, Wehbeh_Rahman_Sharf_2020_iros_nmpc}, they developed centralized and decentralized model predictive control (MPC) methods for the system using a linearized dynamics model. However, several results~\cite{Erunsal_Zheng_Ventura_Martinoli_2022} confirm that linearized MPC underperforms NMPC, especially in highly dynamic tasks aiming to exploit the full system flight envelope. In~\cite{sun2023nmpcmanipulation}, the authors propose a preliminary NMPC method to control the payload's full pose and use the redundancies and achieve inter-robot separation while respecting actuator constraints, but it is only validated in simulation. Moreover, the proposed method in~\cite{sun2023nmpcmanipulation} includes all the cable directions, cable angular velocity, and cable angular jerk in the state vector, making it impractical for scaling to larger numbers of robots. Additionally, the method requires pre-planning the desired trajectory to track the cable states, adding further complexity.

We propose a novel NMPC method for a team of quadrotors to manipulate a rigid-body payload via suspended cables in $6$ DoF. The proposed NMPC exploits the redundancies to achieve additional tasks like inter-robot separation and obstacle avoidance while respecting the full payload dynamics in six degrees of freedom and the actuator constraints. To address real-time computational requirements and scalability, we use a lightweight state vector that only includes payload pose. Additionally, the method only requires planning trajectories for the payload states on the $SE(3)$ manifold, reducing planning complexity. Finally, we present trajectory tracking results in both simulation and real-world experiments to validate our proposed NMPC method.

The paper is organized as follows. In Section~\ref{sec:model}, we review the system's modeling including both payload, cables, and aerial robots evolving on the manifold $SE(3)\times S^2\times SO(3)$. In Section~\ref{sec:nmpc}, we present our novel NMPC method formulation. Section~\ref{sec:experimental_results} presents our experimental results, and Section~\ref{sec:conclusion} concludes the paper.

\begin{figure}[t!]
\centering
  \includegraphics[width=\columnwidth]{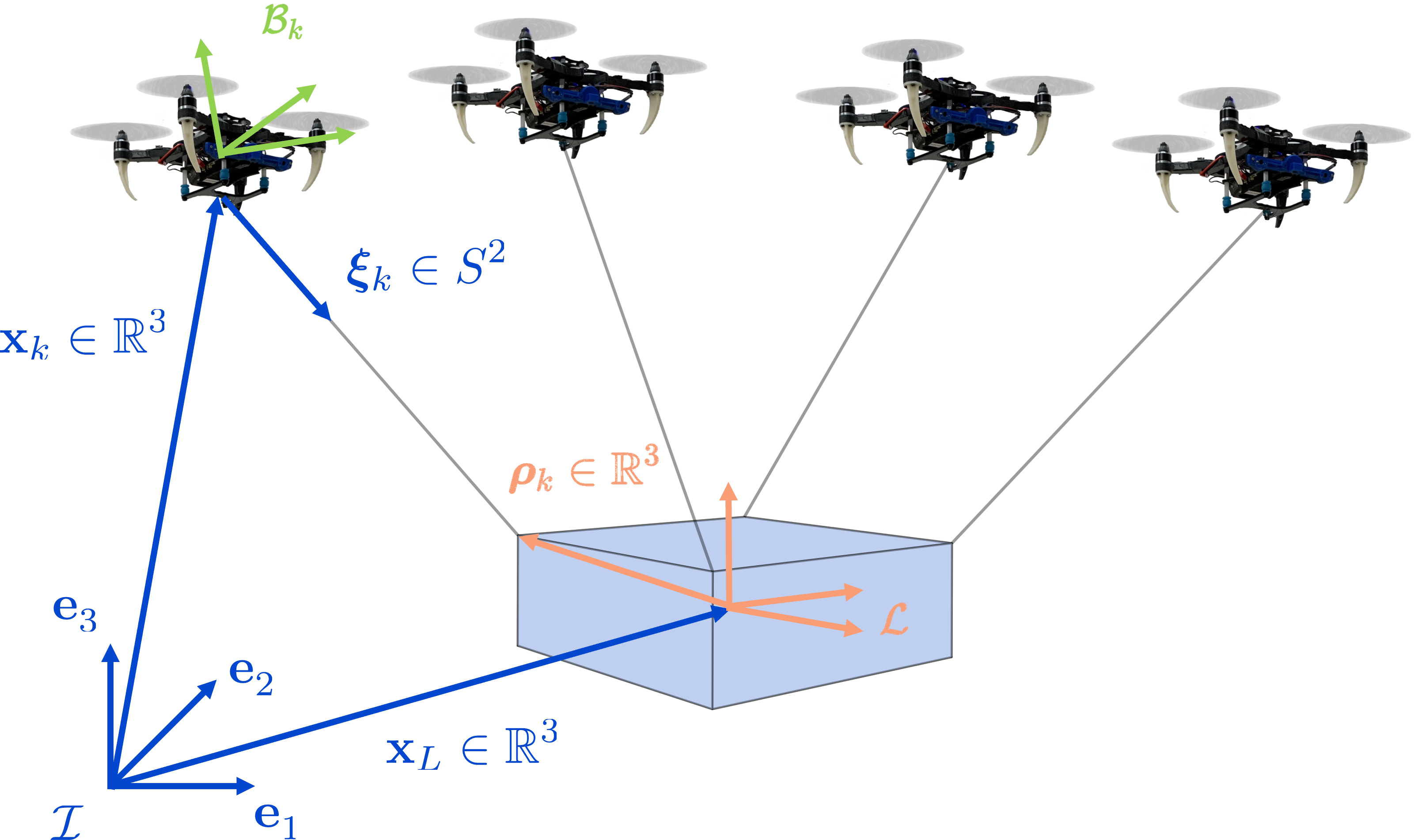}
  \caption{System convention definition: $\worldf$, $\loadf$, $\robotf{\id}$ denote the world frame, the payload body frame, and the $\id^{th}$ robot body frames, respectively, for a generic quadrotor team that is cooperatively transporting and manipulating a cable-suspended payload.}~\label{fig:frame_definition}
\end{figure}

%% file: sections/02-Model_MPC.tex
\begin{table}[t]
\caption {Notation table\label{tab:notation}} 
\centering
\begin{tabularx}{0.48\textwidth}{>{\hsize=0.58\hsize}X >{\hsize=1.42\hsize}X}
    \hline\hline
 $\worldf$, $\loadf$, $\robotf{\id}$ & inertial frame, payload frame, $\id^{th}$ robot frame\\
 $\loadmass,m_{\id}\in\realnum{}$ &  mass of payload, $\id^{th}$ robot\\
 $\loadpos,\robotpos{\id}\in\realnum{3}$ &  position of payload, $\id^{th}$ robot in $\worldf$\\
 $\loadvel, \loadacc\in\realnum{3}$ & linear velocity, acceleration of payload in $\worldf$\\
$\robotvel{\id}, \robotacc{\id}\in\realnum{3}$ &linear velocity, acceleration of $\id^{th}$ robot in $\worldf$\\
 $\loadrot\in\SOthree$&orientation of payload with respect to $\worldf$ \\
 $\robotrot{\id}\in\SOthree$&orientation of $\id^{th}$ robot with respect to $\worldf$ \\
 $\loadangvel$, $\loadangacc\in\realnum{3}$ & angular velocity, acceleration of payload in $\loadf$\\
  $\robotangvel{\id}\in\realnum{3}$& angular velocity of $\id^{th}$ robot in $\robotf{\id}$\\
  $f_{\id}\in\realnum{}$, $\matM_{\id}\in\realnum{3}$&total thrust, moment at $\id^{th}$ robot in $\robotf{\id}$.\\
 $\inertiaload,\inertia_{\id}\in\realnum{3\times3}$   &  moment of inertia of payload, $\id^{th}$ robot\\
 $\cablevec{\id}\in S^2$&unit vector from $\id^{th}$ robot to attach point in $\worldf$\\
 $\bm{\omega}_{\id}\in\realnum{3}$, $l_{\id}\in\realnum{}$&angular velocity, length of $\id^{th}$ cable\\
 $\vecrho{\id}\in\realnum{3}$&position of $\id^{th}$ attach point in $\loadf$\\
    \hline\hline
\end{tabularx}
\end{table}
\section{System Model}\label{sec:model}
In this section, we introduce the system dynamics model for cooperative transportation and manipulation of a rigid body payload by $n$ quadrotors using massless cables. An example of four robots carrying a rigid-body payload is shown in Fig.\ref{fig:frame_definition}. First, we discuss the dynamics of the suspended payload on $SE(3)$ in Section~\ref{sec:payload_dynamics}. Next, we analyze the system redundancy and offer insights for leveraging it in subsequent controller design in Section~\ref{sec:null_space_model}. Lastly, we present the quadrotor's dynamics constrained on the $S^2\times SO(3)$ manifold in Section~\ref{sec:quadrotor_dynamics}. The dynamics model assumes the following
\begin{enumerate}
    \item The drag on the payload and quadrotor is negligible;
    \item The massless cable attaches to the robot's center of mass;
    \item The aerodynamic effect among the robots and the payload is neglected.
\end{enumerate} 
The notation used in this paper is summarized in Table~\ref{tab:notation}.

\subsection{Payload Dynamics}\label{sec:payload_dynamics}
First, let's define the payload state vector, payload velocity vector, and its corresponding derivative
\begin{equation}
    \matX_L = \begin{bmatrix}
        \loadpos\\\loadquat
    \end{bmatrix},\,\matV_L=\begin{bmatrix}
        \loadvel\\\loadangvel
    \end{bmatrix},\,\dot{\matV}_L=\begin{bmatrix}
        \loadacc\\\loadangacc
    \end{bmatrix},
\end{equation}
where $\loadpos,\loadvel,\loadacc$ are the payload's position, velocity, and acceleration with respect to $\worldf$,  $\loadquat$ is the quaternion representation of the payload's orientation with respect to $\worldf$, and $\loadangvel,\loadangacc$ are the payload's angular velocity and acceleration with respect to the payload frame $\loadf$. 
By differentiating both sides of the equation of $\matX_L$, we can obtain the kinematics model of the payload
\begin{equation}
\dot{\matX}_L = \begin{bmatrix}
        \loadvel\\\loadquatdot
    \end{bmatrix} = \begin{bmatrix}
        \loadvel\\\half\loadquat\otimes\loadangvel
    \end{bmatrix} .
    \label{eq:payload_kinematics}
\end{equation}
Further, based on the assumption and classical rigid-body mechanics, the payload's dynamics equation can be easily found as~\cite{RSS2013Koushil}
\begin{equation}
\begin{split}
\loadmass\loadacc{} = \netplforce-\loadmass\vecg,\hspace{0.8em} 
\inertiaload\loadangacc &= \netplmoment-\loadangvel\times\inertiaload\loadangvel,
\label{eq:payload_dynamics}
\end{split}
\end{equation}
where $\netplforce$ is the sum of all the cable tension forces in $\worldf$ and $\netplmoment$ is the total moments generated by the cable forces on the payload in $\loadf$. They can be expressed as
\begin{equation}
\netplforce=\sum_{\id=1}^{n}\tension{\id},\,~\netplmoment=\sum_{\id=1}^{n}\vecrho{\id}\times\loadrot^{\top}\tension{\id},
\label{eq:wrench}
\end{equation}
where $\tension{\id}$ is the $\id^{th}$ cable tension force. By rewriting the payload's equations of motion in matrix form, we can obtain
\begin{equation}
    \dot{\matV}_L  = \bar{\matJ}_L^{-1}\prths{\matW + \matG},
    \label{eq:payload_eq_motion_matrix}
\end{equation}
where $\matW\in\realnum{6}$ denotes the wrench on the payload, $\matG\in\realnum{6}$ is the vector noting the additional physical term like gravity and effect of angular momentum, and $\bar{\matJ}_L\in\realnum{6\times6}$ is the general inertia matrix of the system. They are derived as 
\begin{equation}
\begin{split} 
\matW &= \begin{bmatrix}\netplforce\\\netplmoment\end{bmatrix}, \matG = \begin{bmatrix}-\loadmass\vecg,\\ -\loadangvel\times\inertiaload\loadangvel\end{bmatrix}\\
\bar{\matJ}_L&=\begin{bmatrix}m_L\matI_{3\times3} & \mathbf{0}\\\mathbf{0}& \inertiaload\end{bmatrix}.
    \label{eq:payload_eq_motion_matrix_details}
\end{split}
\end{equation}

\begin{figure}[t!]
\centering
  \includegraphics[width=\columnwidth]{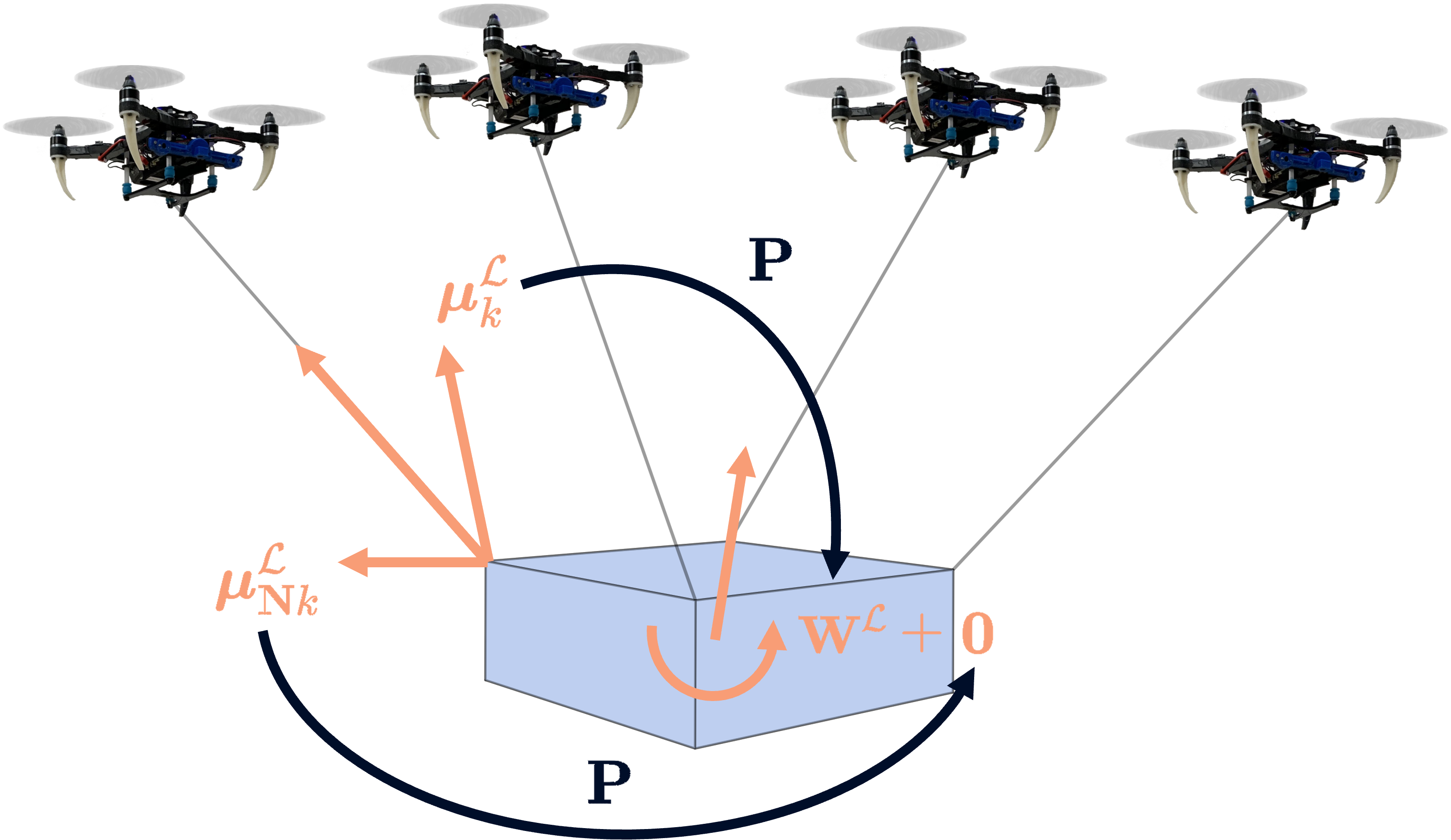}  
  \caption{Visual illustration of the relationship among the payload wrench $\matW^{\loadf}$, cable tension force $\tension{\id}^{\loadf}$ and the tension vectors $\tension{\matN\id}^{\loadf}$ in the null space of the cable tension distribution matrix $\matP$ in the payload frame $\loadf$.\label{fig:null_space_visualization}}
  \vspace{-20pt}
\end{figure}
\subsection{Redundancy in Null Space}\label{sec:null_space_model}
In this section, we analyze the redundancy present in the system when the number of robots $n\geq3$. We provide insights on how to utilize the additional redundancy in controller design to distribute the tension forces for secondary tasks like inter-robot separation and obstacle avoidance.

First, for the convenience of the analysis and controller design later, we express all the forces in the payload frame $\loadf$. Hence the total force on the payload $\netplforce$ and the tension forces $\tension{\id}$ will be rotated into the payload frame $\loadf$ as $\netplforce^{\loadf}$ and $\tension{\id}^{\loadf}$, shown as
\begin{equation}
    \netplforce^{\loadf} = \loadrot^{\top}\netplforce, \hspace{1em}\tension{\id}^{\loadf} = \loadrot^{\top}\tension{\id}.\label{eq:rotate_force_to_load_frame}
\end{equation}
By substituting eq.~(\ref{eq:rotate_force_to_load_frame}) into eq.~(\ref{eq:wrench}) and expressing the result in a matrix form, we can obtain the wrench $\matW^{\loadf}$ in the load frame $\loadf$ as
\begin{equation}
    \matW^{\loadf} = \matP\tension{}^{\loadf},\hspace{1em} \tension{}^{\loadf} = \begin{bmatrix}\tension{1}^{\loadf}\\\vdots\\\tension{n}^{\loadf}\end{bmatrix},
    \label{eq:tension_distribution}
\end{equation}
where $\matP\in\realnum{6\times3n}$ maps tension vectors of all $n$ quadrotors to the wrench on the payload. We derive $\matP$ as
\begin{equation}
\begin{split}
{\matP} =\begin{bmatrix}\matI_{3\times3} & \matI_{3\times3} &\cdots &\matI_{3\times3}\\                              \hatvecrho{1}&\hatvecrho{2}&\cdots&\hatvecrho{n}\end{bmatrix},
\end{split}
\label{eq:P_mat}
\end{equation}
where the hat map $\hat{\cdot} : \realnum{3} \rightarrow \mathfrak{so}(3)$ is defined such that $\hat{\veca}\vecb = \veca \times \vecb, \forall \veca, \vecb \in \realnum{3}$. 

\begin{figure*}[t!]
\centering
  \includegraphics[width=\textwidth]{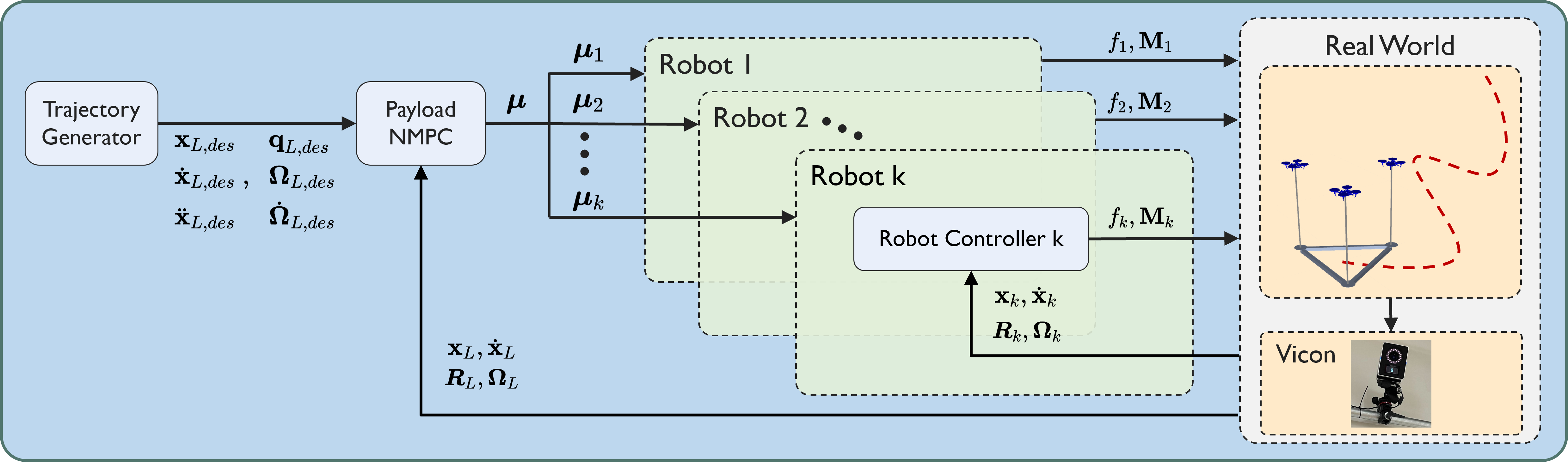}  
  \caption{The control block diagram of the proposed approach.\label{fig:block_diagram}}
  \vspace{-20pt}
\end{figure*}
Upon inspection of $\matP$, we can observe two useful properties. First, all the sub-blocks in the matrix $\matP$ are constant matrices because $\matI_{3\times3}$ is an identity matrix and $\hatvecrho{\id}$ are all skew-symmetric matrix of vector $\vecrho{\id}$ in the $\loadf$, which are all fixed vectors in the $\loadf$. Second, we can ascertain that the matrix $\matP$ exhibits a constant number of rows, independent of the number of robots in the system. For $n \geq 3$ robots, the number of columns of $\matP$ surpasses the number of rows, causing the dimension of the domain of $\matP$ to exceed that of its image. Hence, there is an additional nullity in matrix $\matP$, which the system can utilize to accomplish secondary tasks, such as obstacle avoidance or inter-robot separation~\cite{guanrui2022humanrobot}.

Let $\tension{\matN}^{\loadf} \in \realnum{3n-6}$ be any vector in the null space of matrix $\matP$. Further, we define the basis of the null space of matrix $\matP$ as $\matN \in \realnum{n \times (3n-6)}$, based on the properties of $\matP$ observed previously. And we define $\bm{\Lambda} \in \realnum{3n-6}$ as any coordinate vector with respect to the basis $\matN$. Since the attachment arrangement of the cable mechanism on the payload determines $\matP$ and remains constant, we can obtain $\matN$ for any given setup by knowing $\matP$. Thus, based on the above definition, we can have
\begin{equation}
    \tension{\matN}^{\loadf} = \matN\bm{\Lambda}, ~~\matP\tension{\matN}^{\loadf} = \matP\matN\bm{\Lambda} = 0,
    \label{eq:null_space}
\end{equation}
By adding eq.~\eqref{eq:null_space} into eq.~\eqref{eq:tension_distribution}, we can obtain the following equation
\begin{equation}
    \matW^{\loadf} = \matP\prths{\tension{}^{\loadf}+\matN\bm{\Lambda}}.
    \label{eq:tension_distribution_with_null_space}
\end{equation}
Eq.~\eqref{eq:tension_distribution_with_null_space} implies that by augmenting the tension vector $\tension{}^{\loadf}$ with additional vectors in the null space of matrix $\matP$, it is possible to attain distinct tension distributions that generate an equivalent control wrench on the payload. This highlights the potential to achieve both the desired control wrench $\matW$ on the payload and modify the formation of the robots to meet the demands of secondary tasks through the proper design of $\tension{}$ and $\bm{\Lambda}$.  In Section~\ref{sec:nmpc}, we introduce a novel NMPC method to exploit this property enabling the robots to separate from each other and avoid obstacles in environments meanwhile precisely control the payload's 6-DoF pose. 

\subsection{Quadrotor Dynamics}\label{sec:quadrotor_dynamics}
In this section, we model the $\id^{th}$ quadrotor's dynamics, which is constrained to be on a sphere centered at the corresponding attach point $\id$ as shown in Fig.~\ref{fig:frame_definition}. The location of the $\id^{th}$ quadrotor and the corresponding attach point $\id$ with respect to $\worldf$ is represented by vector $\robotpos{\id}, \attpos{\id}\in\realnum{3}$ respectively. Assuming the cables are all taut, based on the geometric relationship, we can derive that
\begin{equation}
    \robotpos{\id} = \attpos{\id} - l_{\id}\cablevec{\id},
\end{equation}
where $l_{\id}$ is the $k^{th}$ cable's length. The above equation implies the quadrotor's motion is constrained at the surface of a unit sphere centered at the attach point $\id$ as the cable is taut. Based on~\cite{lee2018tcst} and \cite{guanrui2021ral}, we have
 \begin{equation}
 \begin{split}
&\hat{\cablevec{}}_{\id}^2\prths{\inputforce{\id}-m_k\veca_{\id}} = m_{\id} l_{\id}\prths{\cableddotvec{\id} +\twonorm{\cabledotvec{\id}}^2\cablevec{\id}},\\&\veca_{\id}=\loadacc + \vecg-\loadrot\hatvecrho{\id}\loadangacc+\loadrot\hat{\angvel}_L^{2}\vecrho{\id}.
 \end{split}
\label{eq:multi-taut-robot-dyn}
 \end{equation}
 where $\inputforce{k}$ is the input force at the $\id^{th}$ quadrotor, and $\veca_\id$ is the acceleration of the attach point $\id$ with respect to the world frame $\worldf$.

%% file: sections/03-NMPCFormulation.tex
\section{Control}~\label{sec:nmpc}
In this section, we present the hierarchical controller design for the system with $n \geq 3$ quadrotors manipulating a rigid-body payload via cables. The control block diagram is shown in Fig.~\ref{fig:block_diagram}. We first introduce a novel NMPC method in Section~\ref{sec:payloadnmpc} that enables manipulation of the suspended payload pose in all 6 DoFs through cable controlled by $n$ quadrotors. Our proposed NMPC method optimizes the distribution of cable tension forces while exploiting the redundancy in the null space to accomplish secondary tasks, such as obstacle avoidance and inter-robot separation. In Section~\ref{sec:robot_controller}, we then describe a robot controller that takes the first tension vector from the predicted horizon trajectory of the NMPC as input and generates thrust and moment commands to execute the tension force.

\subsection{Payload NMPC}\label{sec:payloadnmpc}
In this section, we propose a novel NMPC method for manipulating the payload's pose. In general, NMPC is a predictive control technique that finds a sequence of system states $\{\matX_{0},\matX_{1},\cdots,\matX_{N}\}$ and inputs, $\{\matU_{0},\matU_{1},\cdots,\matU_{N-1}\}$, over a fixed time horizon of $N$ steps, while optimizing an objective function and respecting nonlinear constraints and system dynamics. The objective function consists of a running cost, $h(\matX,\matU)$ and a terminal cost $h_N(\matX)$ and formulated as
\begin{equation}
    \begin{split}
        \min_{\substack{\matX_{0},\cdots,\matX_{N}, \\ \matU_{0},\cdots,\matU_{N-1}}}& ~\sum_{i=0}^{N-1}  h\left(\matX_{i},\matU_{i}\right) + h_N\prths{\matX_N}\label{eq:MPC_Form},\\ 
        \text{subject to:}&~\matX_{i+1} = \mathbf{f}\left(\matX_{i},\matU_{i}\right),~\forall i = 0,\cdots,N-1\\
        &~\matX{}_0 = \matX{}(t_0),\\
&~\mathbf{g}\left(\matX_{i},\matU_{i}\right)\leq 0,
\end{split}
\end{equation}
where $\matX_{i+1} = \mathbf{f}\prths{\matX_i,\matU_i}$ is the nonlinear system dynamics in discrete form, and  $\mathbf{g}\left(\matX,\matU\right)$ is the additional state and input constraints. The optimization occurs with initial condition $\matX_0$ while respecting system dynamics $\mathbf{f}(\matX, \matU)$. 

In the following, we present our proposed NMPC formulation for transporting a rigid-body payload with $n$ quadrotors using suspended cables. We discuss the advantages of the chosen cost function and describe the nonlinear system dynamics and constraints for exploiting the null space for secondary tasks and respecting actuator limits.
\subsubsection{Cost Function}
First, let us define the state vector and input vector of the NMPC as 
\begin{equation}
    \matX = \begin{bmatrix}\matX_L\\\matV_L\end{bmatrix}, \matU = \begin{bmatrix}\matW\\\veclamb\end{bmatrix}.
\end{equation} 
Since the task is to manipulate the payload to track a desired trajectory in $SE(3)$, we choose the following objective function

\begin{equation}
\begin{split}
    \min_{\matX_i,\matU_i}\,& \vece_{\matX_{N}}^\top\matQ_{X_N}\vece_{\matX_{N}} + \sum_{i=0}^{N-1}\vece_{\matX_{i}}^\top\matQ_X\vece_{\matX_{i}} + \vece_{\matU_{i}}^\top\matQ_U\vece_{\matU_{i}}
    \end{split},\label{eq:nmpc_costs}
\end{equation}
where $\vece_{\matX_{i}}, \vece_{\matU_{i}}$ are the state and wrench errors determined by comparing predicted states and inputs to trajectory references defined as
\begin{equation}
\begin{split}
    \vece_{\matX_{i}} = \mqty[\loadposdes-\loadpos\\\loadveldes -\loadvel\\log_{\mathbf{q}}\prths{\vecq_{des,L}^{-1}\otimes\vecq_{L}}\\\loadangveldes - \loadangvel]_{t=i},~
    \vece_{\matU_{i}} = \mqty[\netplforcedes-\netplforce\\\netplmomentdes-\netplmoment\\\bm{\Lambda}]_{t=i}.
\end{split}
\end{equation}
where $\log_{\mathbf{q}}$ is the logarithm function~\cite{sola2017QuaternionKF}.
Our proposed objective function and NMPC formulation offer several advantages over the NMPC presented in \cite{sun2023nmpcmanipulation}.  Instead of the complex planning required in \cite{sun2023nmpcmanipulation}, which involves planning for the trajectory of the payload states, cable directions, derivatives of cable directions, and cable tension magnitude, the proposed formulation only requires planning for the payload states in $SE(3)$ and their corresponding derivatives, such as velocity and acceleration. Once we obtain the desired payload velocity and acceleration, we can easily derive the corresponding desired wrench $\netplforcedes, \netplmomentdes$ as follows
\begin{equation}
    \netplforcedes = \loadmass\loadacc, \,\, \netplmomentdes = \inertiaload\loadangaccdes + \loadangveldes\times\inertiaload\loadangveldes.
\end{equation}
The optimization dimensionality of our proposed NMPC is approximately four times lower than the approach in \cite{sun2023nmpcmanipulation}, with a variable dimension of $13+3n$ in our NMPC, compared to $13+13n$ in \cite{sun2023nmpcmanipulation}. Furthermore, we include the null space coefficient vector $\bm{\Lambda}$ in the NMPC formulation to exploit the null space to address secondary tasks, such as obstacle avoidance and inter-robot spatial separation, while respecting the nonlinear system dynamics and actuator constraints. 

We solve the proposed NMPC problem in eq.~(\ref{eq:MPC_Form}) with cost as eq.~(\ref{eq:nmpc_costs}) via Sequential Quadratic Programming (SQP) in a real-time iteration (RTI) scheme using ACADOS~\cite{Verschueren2019}. We choose HPIPM as the solver for the quadratic programming in SQP. 
\subsubsection{System Dynamics}
Based on the eqs.~\eqref{eq:payload_kinematics} and \eqref{eq:payload_eq_motion_matrix}, we can obtain the dynamics equation for the state $\matX$ as
\begin{equation}
    \dot{\matX} = \vecf\prths{\matX,\matU}=\begin{bmatrix}
        \loadvel\\\half\loadquat\otimes\loadangvel\\\frac{1}{m_L} \netplforce - \vecg\\\inertiaload^{-1}\prths{\netplmoment-\loadangvel\times\inertiaload\loadangvel}
    \end{bmatrix}.
    \label{eq:discrete_dynamics}
\end{equation}
To incorporate the dynamics equation in the discrete-time formulation, we apply $4^\text{th}$ order Runge-Kutta method to numerically integrate $\dot{\matX}$ over the sampling time $dt$ given the state $\matX_{i}$ and input $\matU_{i}$ as
\begin{equation}
    \matX_{i+1} = \mathbf{f}_{RK4}\prths{\matX_{i},\matU_i,dt}.
\end{equation} 

\subsubsection{Inter-robot Separation Constraints}
As discussed in Section~\ref{sec:null_space_model} and showed in eq.~\eqref{eq:tension_distribution_with_null_space}, through proper design of $\tension{}$ and $\bm{\Lambda}$, we can obtain desired wrench and modify the $\tension{}$ to satisfy some other tasks simultaneously. Inspired by our previous works~\cite{guanrui2022humanrobot}, we propose the following design
\begin{equation}
    \tension{}^{\loadf} = \begin{bmatrix}\tension{1}^{\loadf}\\\vdots\\\tension{n}^{\loadf}\end{bmatrix} = \matP^{\dagger}\matW^{\loadf} + \matN\bm{\Lambda}, \matW^{\loadf}=\begin{bmatrix}
        \loadrot^{\top}\netplforce\\\netplmoment
    \end{bmatrix},\label{eq:tension_distribution_null_space_loadf}
\end{equation}
where $\matP^{\dagger}$ is the pseudo-inverse of $\matP$. Notably, since the cable tension distribution $\matP^{\dagger}\matW^{\loadf}$ is the minimum norm solution. Via minimizing the square norm of $\bm{\Lambda}$, we minimize the total norm of the cable tension vectors which conserves energy for platforms with SWAP constraints like quadrotors.

Since the cable forces are along the cable's direction, we can derive that 
\begin{equation}
    \cablevec{\id}^{\loadf} = -\frac{\tension{\id}^{\loadf}}{\norm{\tension{\id}^{\loadf}}}, \,\,\robotpos{\id}^{\loadf} = \vecrho{\id} - l_\id\cablevec{\id}^{\loadf} = \vecrho{\id} +l_{\id}\frac{\tension{\id}^{\loadf}}{\norm{\tension{\id}^{\loadf}}}, \label{eq:robot_pos_loadf}
\end{equation}
where $\cablevec{\id}^{\loadf},\robotpos{\id}^{\loadf}$ are the cable direction and the robot position with respect to the load frame $\loadf$. By utilizing eqs.~\eqref{eq:tension_distribution_null_space_loadf} and \eqref{eq:robot_pos_loadf}, we can construct constraints on the robot position to guarantee inter-robot spatial separation as
\begin{equation}
    \left\|\robotpos{k}^{\loadf}-\robotpos{j}^{\loadf}\right\|^2 \geq d_r^2,  \,\,\,\, 0<k<j\leq n,\\
\end{equation}
where $d_r$ is the minimum distance between robots that we can choose. 
\subsubsection{Obstacle Avoidance Constraints}
Let the obstacle position with respect to the $\worldf$ be $\vecp_{O}$. We can define the following constraints in the NMPC such that the robot team and the payload can keep a safe distance from obstacles in the world frame $\worldf$
\begin{equation}
 \|\vecp_{O}-\loadpos-\loadrot\robotpos{k}^{\loadf}\|^2 \geq d_{Or}^2,
\end{equation}
\begin{equation}
 \|\vecp_{O}-\loadpos\|^2 \geq d_{OL}^2.
\end{equation}
\subsubsection{Actuator Constraints}
As we obtain the predicted cable tension force from the eq.~\eqref{eq:tension_distribution_null_space_loadf}, we can further limit the tension force norm to provide some boundary for the actuators
\begin{equation}
 \norm{\vecmu_{\id}^{\loadf}} \leq f_{max}.
\end{equation}

\begin{figure*}
    \centering
    \includegraphics[width=\textwidth]{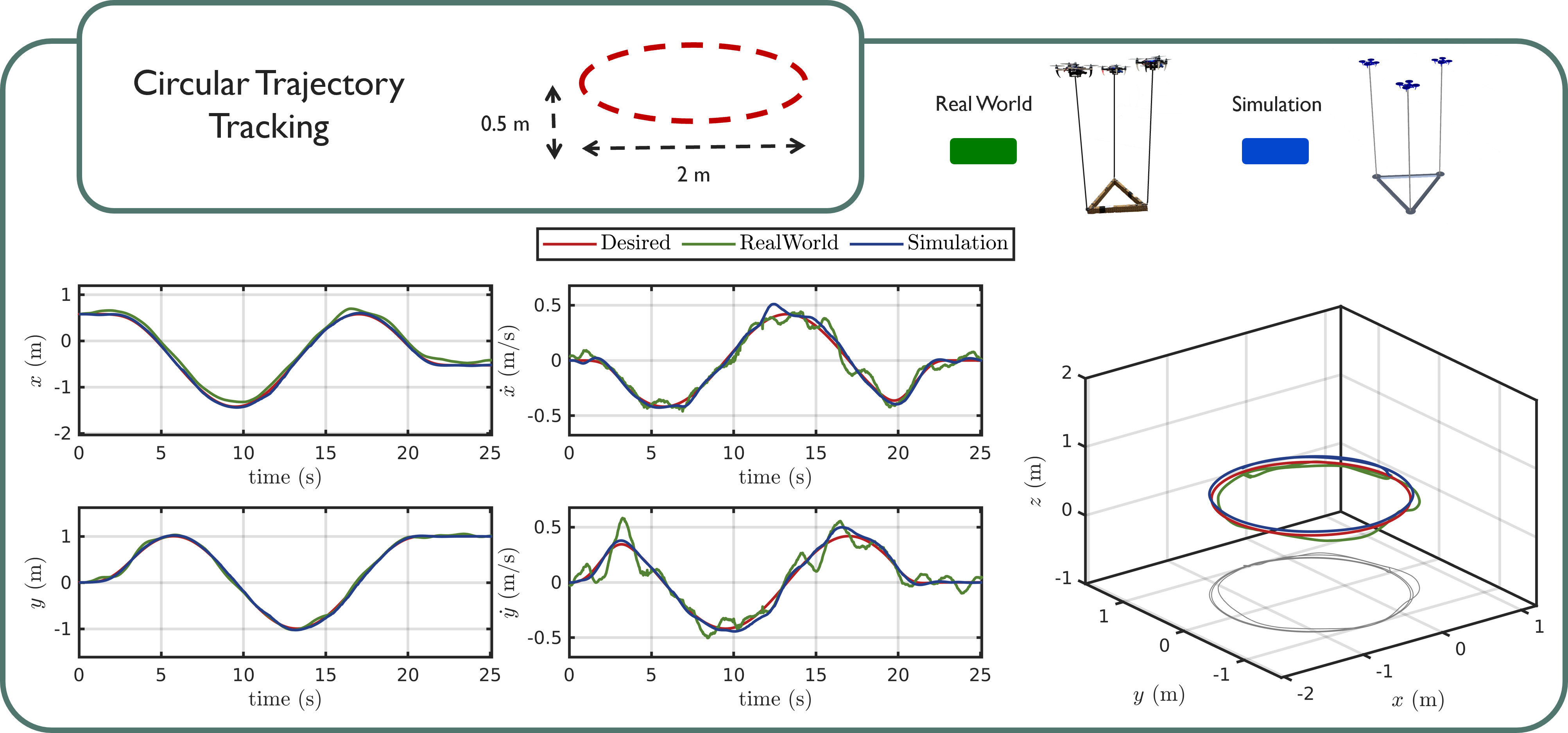}
    \caption{Tracking results of the payload following a circular trajectory in x-y plane.}
    \label{fig:circular_traj}
    \vspace{-10pt}
\end{figure*}
\subsection{Robot Controller}\label{sec:robot_controller}
As we solved the optimization problem in eq.~\eqref{eq:MPC_Form}, we obtain a sequence of system states $\{\matX_{0},\matX_{1},\cdots,\matX_{N}\}$ and inputs, $\{\matU_{0},\matU_{1},\cdots,\matU_{N-1}\}$, over a fixed time horizon of $N$ steps. 
We choose the first input $\matU_{0}= \begin{bmatrix}\matW_0\\\veclamb_0\end{bmatrix}$ in the input sequence to obtain the desired tension force for the $k^{th}$ quadrotor 
\begin{equation}
    \begin{bmatrix}\tension{1,des}\\\vdots\\\tension{n,des}\end{bmatrix} = \text{diag}\prths{{\loadrot,\cdots,\loadrot}}\prths{ \matP^{\dagger}\begin{bmatrix}
        \loadrot^{\top}\netplforce_0\\\netplmoment_0
    \end{bmatrix} + \matN\bm{\Lambda}_0}.
\end{equation}
Once obtained the desired tension forces $\tension{des}$, the tension input of each cable of the individual quadrotor is selected as the projection of desired tension on the corresponding cable 
 \begin{equation}
     \tension{\id}= \cablevec{\id}\cablevec{\id}^{\top}\tensiondes{\id},
 \end{equation}
The desired direction $\cablevec{\id,des}$ and the desired angular velocity $\cableveldes{\id}$ of the $\id^{th}$ cable link can be obtained as
\begin{equation*}
    \cablevec{\id,des} = -\frac{\tensiondes{\id}}{\norm{\tensiondes{\id}}},\,\cableveldes{\id} = \cablevec{\id,des}\times\cabledotvec{\id,des},
\end{equation*} 
where $\cabledotvec{\id,des}$ is the derivative of $\cablevec{\id,des}$. The thrust $f_{\id}$ and moments $\mathbf{M}_{\id}$ acting at the $\id^{th}$ quadrotor are
\begin{equation}
    f_{\id} = \inputforce{\id}\cdot\robotrot{\id}\axis{3}{} = \prths{\inputpara{\id} + \inputperp{\id}}\cdot\robotrot{\id}\axis{3}{}, \label{eq:ctrl_force_scalar}
\end{equation}
\begin{align}
\label{eq:ctrl_moment} 
\matM_{\id} =& \mathbf{K}_{R}\mathbf{e}_{R_{\id}} + \mathbf{K}_{\bm{\Omega}}\mathbf{e}_{\bm{\Omega}_{\id}} + \bm{\Omega}_{\id}\times\inertia_{\id}\bm{\Omega}_{\id}\\
&-\inertia_{\id}\prths{\hat{\bm{\Omega}}_{\id}\robotrot{\id}^{\top}\robotrot{\id,des}\bm{\Omega}_{\id,des}-\robotrot{\id}^{\top}\robotrot{\id,des}\dot{\bm{\Omega}}_{\id,des}},\nonumber
\end{align}
where $\matK_{R}$ and $\matK_{\bm{\Omega}}$ are diagonal positive constant matrices, $\mathbf{e}_{R_{\id}}$ and $\mathbf{e}_{\bm{\Omega}_{\id}}$ are the orientation and angular velocity errors defined as follows
\begin{equation}
\begin{split}
\vece_{\robotrot{\id}} &= \half\prths{\robotrot{\id}^{\top}\robotrot{\id,des}-\robotrot{\id,des}^{\top}\robotrot{\id}}^{\vee}, \\
\vece_{\robotangvel{\id}} &=  \loadrot^{\top}\loadrotdes\loadangveldes - \loadangvel.\label{eq:control_errors}
\end{split}
\end{equation}
The inputs $\inputperp{\id}$ and $\inputpara{\id}$ are designed as
\begin{equation}
\begin{split}
    \inputperp{\id}  =& m_{\id} l_{\id}\cablevec{\id}\times\left[-\matK_{\cablevec{\id}}\vece_{\cablevec{\id}} -\matK_{\cablevel{\id}}\vece_{\cablevel{\id}} -\hat{\cablevec{}}_{\id}^2\cableveldes{\id}\right.\\
    &\left.-\prths{\cablevec{\id}\cdot\cableveldes{\id}}\cabledotvec{\id,des}\right]- m_{\id}\hat{\cablevec{}}_{\id}^2\veca_{\id,c},\\
\end{split}
\end{equation}
\begin{equation}
\inputpara{\id}  = \tension{\id} + m_\id l_{\id}\twonorm{\cablevel{\id}}^2\cablevec{\id}  + m_\id\cablevec{\id}\cablevec{\id}^{\top}\veca_{\id,c},
\end{equation}
where
 \begin{equation}
    \veca_{\mathit{\id},c} = \loadaccdes + \vecg -\loadrot\hatvecrho{\id}\loadangacc+ \loadrot\hat{\angvel}_{L}^2\vecrho{\id},
    \label{eq:control_inputs}
\end{equation}
where $\matK_{\cablevec{\id}}$ and $\matK_{\cablevel{\id}}$ are diagonal positive constant matrices, $\vece_{\cablevec{\id}}$ and $\vece_{\cablevel{\id}}$ are the cable direction and cable angular velocity errors respectively
\begin{equation*}
\vece_{\cablevec{\id}}= \cablevec{\id,des} \times\cablevec{\id},~
\vece_{\cablevel{\id}} = \cablevel{\id} + \cablevec{\id}\times
\cablevec{\id}\times\cableveldes{\id}.
\end{equation*}
Readers can refer to~\cite{Leecdc2014} for a stability analysis for the robot controller.

%% file: sections/04-Experiments.tex
\begin{figure*}
    \centering
    \includegraphics[width=\textwidth]{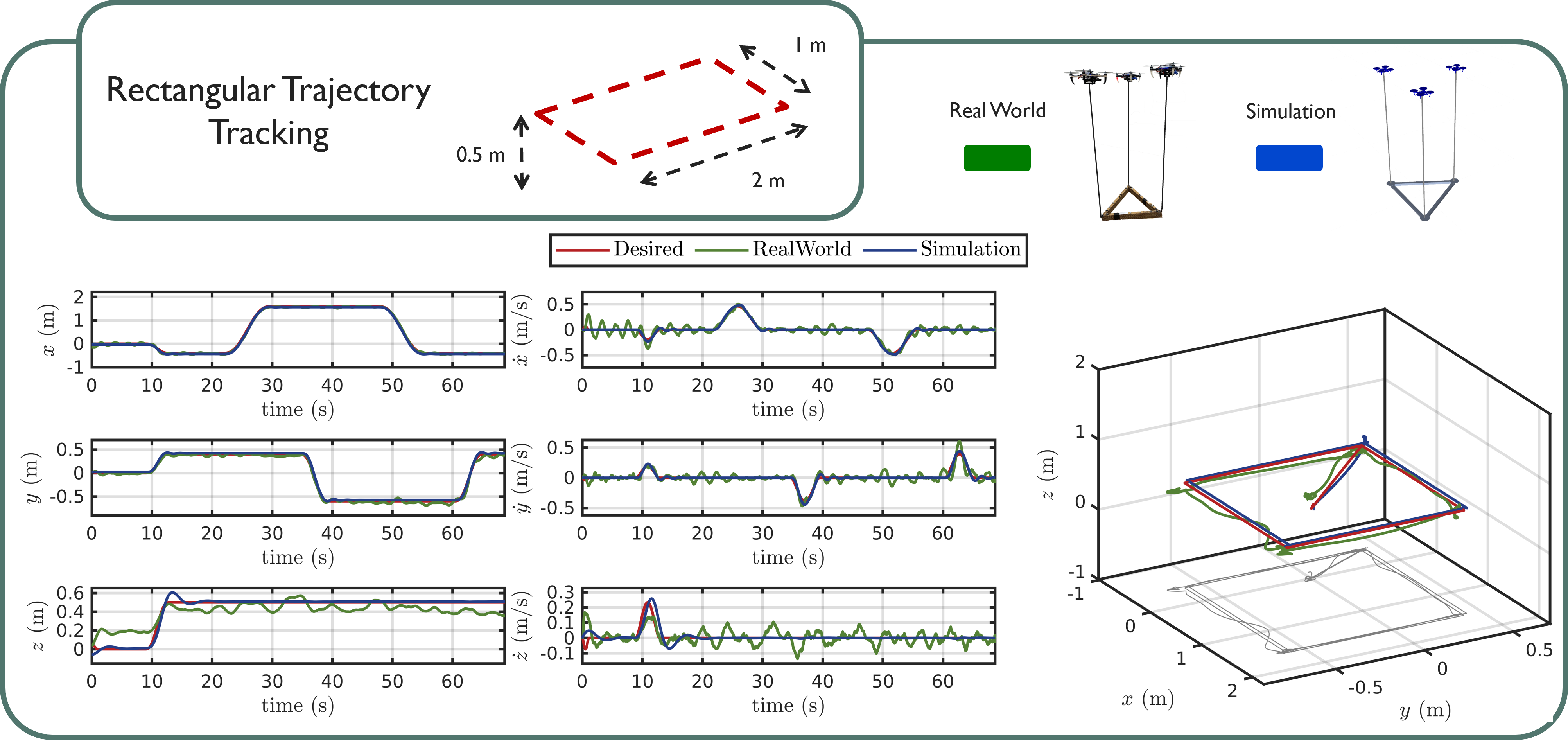}
    \caption{Tracking results of the payload taking off from the ground and following a rectangular trajectory in x-y plane.}
    \label{fig:rectangular_traj}
    \vspace{-10pt}
\end{figure*}
\section{Experiments}~\label{sec:experimental_results}
In this section, we validate our NMPC method with experimental results in both simulation and real-world settings. We introduce our experimental setup, including environments and platforms, and showcase the results of three quadrotors transporting a rigid-body payload through rectangular and circular trajectories. Additional experiments like obstacle avoidance are shown in our multimedia materials.
\subsection{Setup}
We conduct the simulation experiments using our open-source simulator~\cite{guanrui2022rotortm}, in which we tuned the NMPC parameters, including the horizon and weights. We then apply the same controller parameters to real-world experiments.
The real-world experiments were performed in the $10\times6\times4~\si{m^3}$ flying space of the ARPL lab at New York University, using a triangular payload suspended from three quadrotors, each connected by a cable of length $l_k=1~\si{m}$. The quadrotor platform was equipped with a $\text{Qualcomm}^{\circledR}\text{Snapdragon}^{\text{TM}}$ board for on-board computing~\cite{LoiannoRAL2017}, and the system was developed using ROS\footnote{\url{www.ros.org}}. The payload mass, at $232~\si{g}$, exceeded the capacity of each single vehicle and our previous system~\cite{LoiannoRAL2018}. We estimated the odometry of the payload and quadrotors, and the position and velocity of attachment points using a Vicon\footnote{\url{www.vicon.com}} motion capture system at $100~\si{Hz}$. The unit vector of each cable direction $\cablevec{k}$ and the corresponding velocity $\cabledotvec{k}$ were estimated by
\begin{equation}
    \cablevec{k} = \frac{\aptpos{k} - \robotpos{k}}{\norm{\aptpos{k} - \robotpos{k}}},~~~\cabledotvec{k} = \frac{\aptvel{k} - \robotvel{k}}{l_k},
\end{equation}
where $\aptpos{k}, \aptvel{k}$ are the position and velocity of the $k^{th}$ attach point in $\worldf$ and $\robotpos{k},\robotvel{k}$ are position and velocity of the $k^{th}$ robot in $\worldf$ , all of which are estimated by the motion capture system.

\begin{table}[!t]
\caption {Payload trajectory tracking error and RSME.\label{tab:tracking_RMSE}} 
\centering
\begin{tabularx}{0.48\textwidth}{>{\hsize=0.5\hsize}X >{\hsize=1.42\hsize}X>{\hsize=0.58\hsize}X >{\hsize=1.42\hsize}X>{\hsize=0.58\hsize}X >{\hsize=1.42\hsize}X}
    \hline\hline
 \rule{0pt}{2ex} &Component&\multicolumn{2}{c}{Circular}&\multicolumn{2}{c}{Rectangular}\\
  \cline{3-6}\
  \rule{0pt}{2ex}  &   & Real   &  Sim   & Real & Sim\\\hline
  Position&x$~(\si{m})$&0.03800 & 0.03574 & 0.08766  & 0.07998 \\
          &y$~(\si{m})$&0.03610 & 0.02665 & 0.03883  & 0.02101 \\
          &z$~(\si{m})$&0.09714 & 0.02245 & 0.08868  & 0.09806 \\
    \hline\hline
\end{tabularx}
\vspace{-10pt}
\end{table}
\subsection{Trajectory Tracking}
In this section, we evaluate our proposed NMPC performance in trajectory tracking experiments. The results are shown in Figs.~\ref{fig:circular_traj} and \ref{fig:rectangular_traj}.
Two types of trajectories are executed, a circular trajectory and a rectangular trajectory. 
The circular trajectory is designed as 
\begin{equation*}
  \loadposdes^{c}(t) = \begin{bmatrix}r\cos\frac{2\pi t}{T_c}&r\sin\frac{2\pi t}{T_c}&h\end{bmatrix}^\top,
\end{equation*}
with period $T_c = 15~\si{s}$, radius $r = 1.0~\si{m}$ and a constant height $h = 0.5~\si{m}$. The rectangular trajectory is designed length of $2~\si{m}$ in x direction and $1~\si{m}$ in the y direction.
In Figs.~\ref{fig:circular_traj} and \ref{fig:rectangular_traj}, we can observe that the position and velocity of the center of mass of the payload closely follow the commanded trajectory.  Table~\ref{tab:tracking_RMSE} reports the RMSE of the payload's position compared to the desired trajectory, which is within a range of $0.1~\si{m}$. These results demonstrate that our proposed NMPC method can transport the payload with small tracking errors.
\begin{figure}[!t]
    \centering
    \includegraphics[width=\columnwidth]{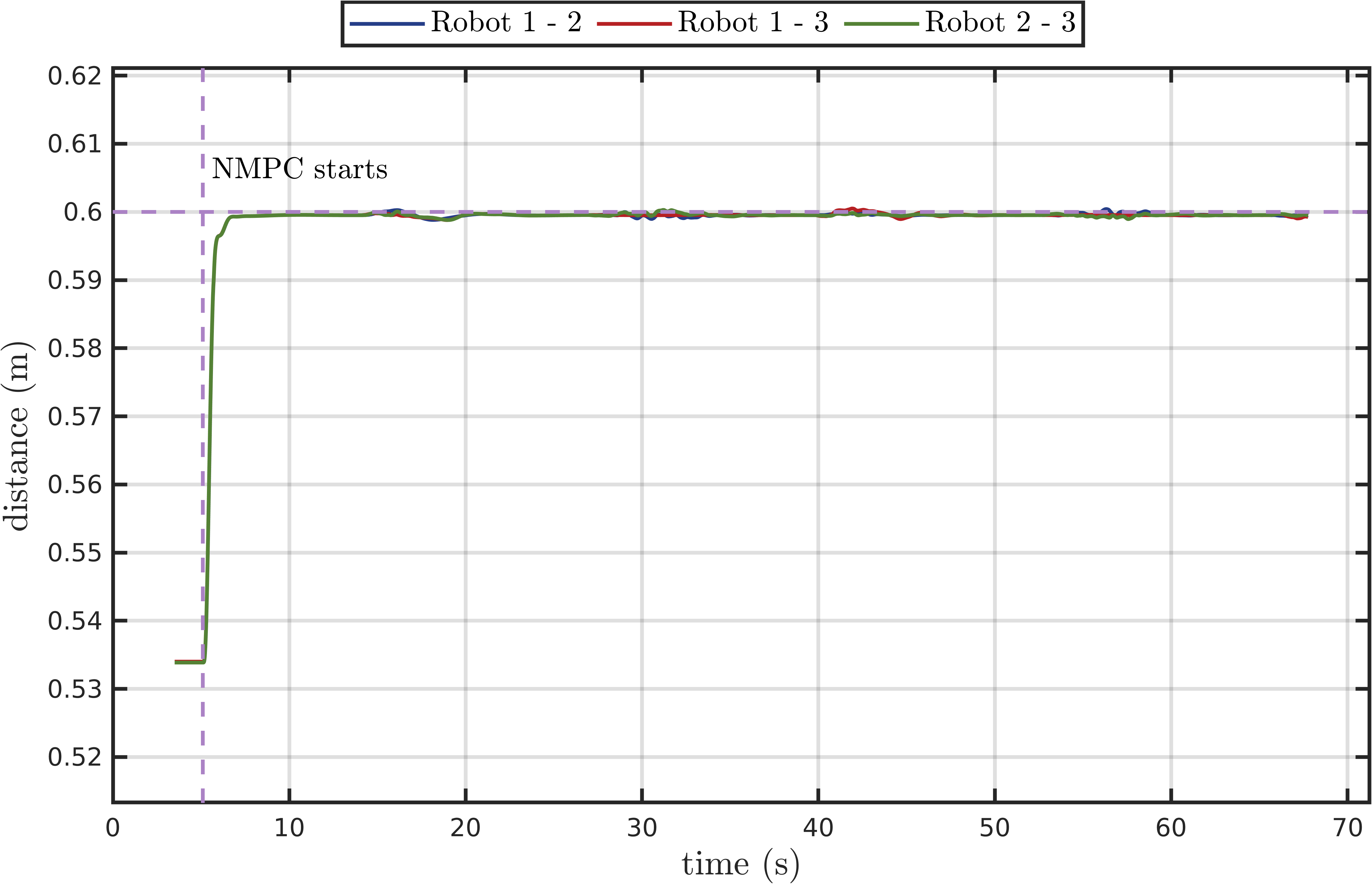}
    \caption{Inter-robot distance in the rectangular trajectory tracking in simulation, staying at the boundary of $0.6~\si{m}$}.
    \label{fig:robot_distance}
    \vspace{-20pt}
\end{figure}
We also present the results of our method's ability to exploit the null space for inter-robot spatial separation. Fig.~\ref{fig:robot_distance} shows the inter-robot distance results in the rectangular trajectory tracking experiment in simulation. Before the NMPC starts running, the inter-robot distances are around $0.53~\si{m}$, which is smaller than the boundary of $0.6~\si{m}$ that we set in the parameters. After the NMPC starts, the inter-robot distance starts to grow and reaches $0.6~\si{m}$. All the inter-robot distance stays at the boundary as we optimize to minimize the null space coefficients square norm for saving energy. Further, the tension constraints can also implicitly guarantee that the actuator constraints are satisfied.

%% file: sections/05-Conclusion.tex
\section{Conclusion}~\label{sec:conclusion}
In this paper, we presented a novel NMPC method that enables a team of quadrotors to manipulate a rigid-body payload in all 6 degrees of freedom via suspended cables. Our approach exploits redundancies to perform additional tasks like inter-robot separation and obstacle avoidance while respecting payload dynamics and actuator constraints. We validated the NMPC method with trajectory tracking experiments in both simulation and real-world scenarios. The results show that the proposed NMPC can manipulate the rigid-body payload with small tracking errors meanwhile satisfying the additional constraints.

 Future works will investigate the ability to complement the proposed formulation with perception objectives or constraints to fully exploit the power of the proposed method in case of autonomous tasks. In addition, there are still some open research questions that would be interesting to investigate such as the robustness of the control algorithms to uncertainties, such as external disturbances, measurement noise, scalability with respect to the number of robots and load sizes, and changes in the system parameters. A clear understanding of these aspects can contribute to guarantee reliable operation of the system in real-world scenarios. 